\pdfoutput=1

\documentclass[11pt]{article}

\usepackage[final]{EMNLP2023}

\usepackage{times}
\usepackage{latexsym}
\usepackage{amsfonts}
\usepackage{amsmath}
\usepackage{booktabs}
\usepackage{multirow}
\usepackage{amssymb}
\usepackage{pifont}

\newcommand{\cmark}{\ding{51}}
\newcommand{\xmark}{\ding{55}}

\usepackage[T1]{fontenc}

\usepackage[utf8]{inputenc}

\usepackage{microtype}

\usepackage{inconsolata}

\usepackage{graphicx}

\title{Empowering Large Language Model for Continual Video Question Answering with Collaborative Prompting}

\author{Chen Cai\thanks{~Equal contributions.} , Zheng Wang\footnotemark[1] , Jianjun Gao, Wenyang Liu \\ \textbf{Ye Lu, Runzhong Zhang, Kim-Hui Yap\thanks{~Corresponding author.}}\\
  Nanyang Technological University \\
  \texttt{\{e190210, zheng011, ekhyap\}@ntu.edu.sg}}

\begin{document}
\maketitle
\begin{abstract}
In recent years, the rapid increase in online video content has underscored the limitations of static Video Question Answering (VideoQA) models trained on fixed datasets, as they struggle to adapt to new questions or tasks posed by newly available content. 
In this paper, we explore the novel challenge of VideoQA within a continual learning framework, and empirically identify a critical issue: fine-tuning a large language model (LLM) for a sequence of tasks often results in catastrophic forgetting. To address this, we propose Collaborative Prompting (ColPro), which integrates specific question constraint prompting, knowledge acquisition prompting, and visual temporal awareness prompting. These prompts aim to capture textual question context, visual content, and video temporal dynamics in VideoQA, a perspective underexplored in prior research. Experimental results on the NExT-QA and DramaQA datasets show that ColPro achieves superior performance compared to existing approaches, achieving 55.14\% accuracy on NExT-QA and 71.24\% accuracy on DramaQA, highlighting its practical relevance and effectiveness. Code available at \url{https://github.com/caicch/ColPro}
\end{abstract}

\section{Introduction}

\begin{figure}[t]
\centering
  \centerline{\includegraphics[width=8.3cm]{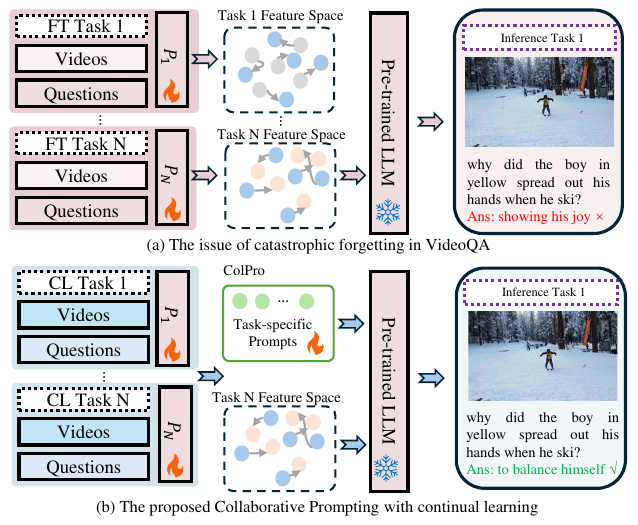}}
  \caption{(a) Existing fine-tuning techniques train for different VideoQA tasks, which could lead to catastrophic forgetting, and generate inferior results. (b) We introduce the Collaborative Prompting (ColPro) within a continual learning framework, which retains task-specific knowledge to generate accurate answers, where $\mathbf{P}_N$ denotes a projection layer.}
  \label{fig:illustration}
\end{figure}

Video Question Answering (VideoQA) is critical for video understanding, involving training machine learning models to accurately respond to questions across various tasks (e.g., finding specific information~\citealp{drama-qa}, counting objects~\citealp{Next-qa}, recalling actions~\citealp{Action-qa}) based on given video content. However, existing VideoQA models are typically trained on fixed datasets in static environments.
With a continual increase in the number of videos on the internet every day, these static models may face challenges in answering new questions posed by the newly available content.
One straightforward solution to overcome this challenge is to fine-tune the models when new data is introduced. However, this approach can lead to higher computational costs when retraining on all the data. Alternatively, fine-tuning only on newly available video question-and-answer pairs may lead to the catastrophic forgetting issue~\cite{catastrophic}, as shown in Figure~\ref{fig:illustration}(a). 

This motivates us to explore continual learning techniques~\cite{icarl, replayCL, l2p} for VideoQA, facilitating ongoing fine-tuning of models across a sequence of data while mitigating catastrophic forgetting of previous tasks (e.g., finding information, or counting objects mentioned earlier), thereby addressing the needs of real-world dynamic environments.
Recent continual learning techniques~\cite{l2p, Dualprompt} have achieved good performance by employing rehearsal-free methods, such as learnable prompting~\cite{Visual_prompt} and prefix-tuning ~\cite{Prefix-tuning}. These approaches eliminate the need for memory-intensive stored experiences from previous tasks~\cite{replayCL, Co2l}, reduce computation costs, and minimize forgetting. 
Specifically, L2P~\cite{l2p}, DualPrompt~\cite{Dualprompt}, DBI~\cite{Decouple_vqa}, and ProgPrompt \cite{progprompt} employ task-aware prompting techniques to fine-tune pre-trained models for downstream tasks using fewer learnable parameters. While these methods have improved performance in vision and language continual learning tasks, they often transfer either single-modal (text only) or multimodal (text and images) information from task to task.
In terms of VideoQA, it is crucial to incorporate textual question context (\textbf{A1}), visual content (\textbf{A2}), and video temporal dynamics (\textbf{A3}) in the continual training setting. In this paper, we introduce Collaborative Prompting (ColPro), which explores these aspects for the VideoQA problem, and represents an area that has not been fully explored in prior research, as shown in Figure~\ref{fig:illustration}(b).

The core idea of ColPro is to empower a base model to achieve enhanced performance when transferring across a sequence of tasks. Inspired by the robust reasoning abilities of recent Large Language Models (LLMs), we instantiate the base model as a LLM (e.g., LLaMA~\citealp{llama}) to generate accurate answers from textual questions and video inputs. Specifically, ColPro integrates three types of prompting techniques: task-specific question constraint prompting (TQCP), knowledge acquisition prompting (KAP), and visual temporal awareness prompting (VTAP), aimed at enhancing accuracy in answer prediction while minimizing forgetting.
TQCP enables the model to gain awareness of the task type and select the correct prompt representation using a negative guiding approach~\cite{NegGuided, negpro}. This method allows the prompt representation to positively correlate with the current task-specific question and negatively correlate with negative question samples. Additionally, KAP acquires task-specific question and video information to enhance accurate answer prediction (for \textbf{A1}). Furthermore, VTAP integrates visual information with the LLM by continuously incorporating video dynamics into prompts through autoregressive temporal dynamics and video distillation loss (for \textbf{A2} and \textbf{A3}). With these prompting strategies, ColPro encapsulates multimodal information to enhance task-specific question answering and mitigate catastrophic forgetting during inference.

Our main contributions to this paper are as follows: (1) We explore the novel problem of video question answering (VideoQA) in a continual learning context, and demonstrate a critical issue: efficiently fine-tuning a LLM for a sequence of tasks leads to catastrophic forgetting. This motivates us to conduct empirical studies to mitigate this issue. 
(2) We propose Collaborative Prompting (ColPro), which utilizes three distinct aspects: textual question context, visual content, and video temporal dynamics, in VideoQA to facilitate knowledge transfer to future tasks.
(3) We conduct extensive experiments on the split VideoQA dataset (NExT-QA~\cite{Next-qa} and DramaQA~\cite{drama-qa}) for continual task-specific answer prediction. Our findings show that ColPro achieves state-of-the-art results, with 55.14\% accuracy on NExT-QA and 71.24\% accuracy on DramaQA.

\section{Related Works}
\subsection{Video Question Answering}
VideoQA is a fundamental task in video understanding, aiming to answer questions based on video content~\cite{ContrasQA, MIST, drama-qa}. Many recent works have explored LLM-based VideoQA~\cite{Self-chained-VQA, Valley, causalVQA}, which requires a LLM to predict the correct answer given a video and question pair. Flipped-VQA~\cite{causalVQA} uses the prompting technique to fine-tune a LLM to learn the specific VideoQA task. SeViLA~\cite{Self-chained-VQA} is built based on a pre-trained large image-language model~\cite{blip2}, extending its capabilities to perform reasoning on video inputs. However, most existing methods are trained on fixed datasets to handle reasoning in static environments, which struggle to answer new questions or tasks posed by newly available content. In contrast, we study a continual video question answering problem, and address its inherent challenges caused by catastrophic forgetting.

\begin{figure*}[t]
 \centerline{\includegraphics[width=16cm]{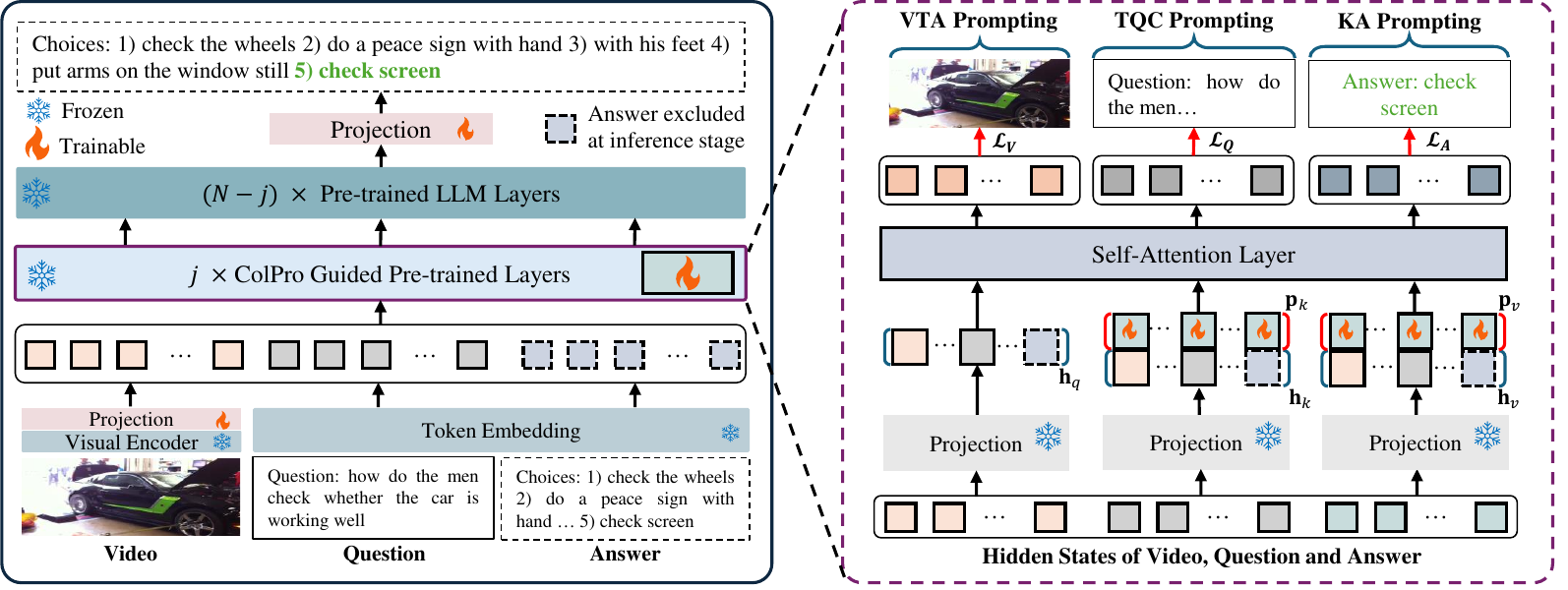}}
  \caption {Illustration of the Collaborative Prompting (ColPro) framework. \textbf{Left}: The training process incorporates ColPro into the first $j$ ColPro Guided Pre-trained Layers to enhance answer prediction accuracy while minimizing forgetting. \textbf{Right}: Three detailed prompting techniques within ColPro are demonstrated: task-specific question constraint prompting (TQCP), knowledge acquisition prompting (KAP), and visual temporal awareness prompting (VTAP). Together, these techniques encapsulate the textual question context, visual content, and video temporal dynamics for each VideoQA task.}
  \label{fig:overview} 
\end{figure*}

\subsection{Continual Learning for Visual Question Answering}
Over the past few years, various continual learning approaches have been proposed to address the issue of catastrophic forgetting~\cite{LearningWF, icarl, replayCL}. Existing methods can be summarized into three categories: rehearsal-based~\cite{DrakCL, replayCL}, architecture-based~\cite{LtoGrow, CL_sim_dsim}, and regularization-based~\cite{memoryCL, plasticityCL}. Rehearsal-based approaches involve constructing a subset of learned samples in a memory buffer and replaying them when learning a new task. Architecture-based approaches allocate separate sets of dedicated parameters for each different task. Regularization-based approaches preserve changes to weights associated with older tasks and selectively stabilize parameter changes. 
%
Recent studies~\cite{l2p, Dualprompt, LAE} draw inspiration from learnable prompting~\cite{prompting, Llama-adapter} in natural language processing to address catastrophic forgetting by learning a small number parameter that is attached to a pre-trained model. Specifically, L2P~\cite{l2p} utilizes a set of task-specific learnable prompts to dynamically instruct a pre-trained model for continual learning. ProgPrompt~\cite{progprompt} adopts progressive networks with a pre-trained language model to learn prompts for different tasks and sequentially concatenates the task-specific learned prompts for text classification. Learning-Accumulation-Ensemble (LAE)~\cite{LAE} utilizes different Parameter-Efficient Fine-Tuning (PEFT) methods such as adaptor~\cite{Adopter}, Lora~\cite{lora}, and prompting~\cite{prompting} for image classification. 

Recent visual question answering models~\cite{Symbolic_Replay, Decouple_vqa} have been exploring the continual learning techniques to answer new questions with given images without experiencing catastrophic forgetting.~\citealp{NVQACL} and~\citealp{Symbolic_Replay} introduce replay-based method to address image-based question answering tasks.~\citealp{Decouple_vqa} use multimodal decoupled prompts to interact with a pre-trained vision-language model, capturing the intricate relationships between modalities. Similar to adapter-based LAE~\cite{LAE}, Dynamic Adapter Merging (DAM)~\cite{DAM} utilizes an adaptor-based framework~\cite{Adopter} for video question answering. 
Unlike DAM, which addresses domain shift in datasets using an adaptor, our work aims to guide a LLM to comprehend multimodal information, including question context, visual content, and temporal dynamics, through a novel prompting technique called ColPro. To the best of our knowledge, this approach is the first of its kind.

\section{Methodology}

\subsection{Motivation and Objective}
In this section, we provide an overview of our approach. First, we discuss continual learning scenarios and their applications for VideoQA. Next, we explain our motivation for utilizing prompting strategies with LLM to achieve our goals. Finally, we present the overall architecture of the proposed method and its training objective.

\smallskip
\noindent\textbf{Continual Learning Scenarios.} In continual learning scenarios, a model is trained sequentially through various stages using a dataset $D$ = $<$$d_1, d_2, ..., d_T$$>$, where $d_t$ $(1 \le t \le T)$ denotes the $t$-th training task, and data from previous tasks is not accessible during the training of stage $t$. In this paper, we study the problem of rehearsal-free continual learning on video question answering tasks, where the data $d_t$ = $<$$V^t, Q^t, A^t$$>$ consists of video $V^t$, question $ Q^t$, and answer $A^t$ pairs. For our experiments, we segment the types of questions into $T$ tasks followed by~\cite{NVQACL, Symbolic_Replay} to benchmark our proposed approach on the NExT-QA \cite{Next-qa} and DramaQA \cite{drama-qa} datasets. Following the settings in existing rehearsal-free works~\cite{l2p, progprompt}, we assume a pre-trained LLM model (e.g., LLaMA~\cite{llama}) is available for our experiments.

\smallskip
\noindent\textbf{Prompting for LLM-based Video Question Answering.} Prompting, a learnable prompt-based learning technique~\cite{Llama-adapter} has been introduced as a streamlined fine-tuning approach, transforming large language models (e.g., LLaMA \cite{llama}) into highly efficient instruction-following models. The core concept of prompting is to incorporate additional instructions into pre-trained LLMs, enabling them to perform downstream tasks in both NLP and multimodal reasoning contexts~\cite{LLava, Minigpt-4}. In this work, we leverage the efficient instruction-following capabilities and outstanding reasoning abilities of LLMs to achieve accurate multimodal question answering in a continual learning scenario. We illustrate the prompting for LLM-based continual VideoQA as follows.

Our primary focus is on leveraging LLMs for continual learning in VideoQA tasks, with LLaMA-Adapter~\cite{Llama-adapter} serving as our baseline. We adopt their prompt tuning adaptation approach to incorporate task-specific information by learning through LLaMA layers, given the input of task-specific questions, videos, and answers. In the inference stage, the frozen model employs the learned task-specific prompt knowledge to predict task-specific answers. In our framework, given the $N$-layers LLaMA, we inject prompts for the first $j$-layers LLaMA transformer layers, named ColPro Guided Pre-trained Layers $\theta(.)$.
We maintain the pre-trained model frozen while tuning a select few additional learnable prompts. Rather than appending prompts directly to the input tokens, our approach involves adding prompts to the keys and values within the Multihead Self-Attention (MSA) layer, following the structure described in Transformer architectures~\cite{trans}.
With the split sets of learnable prompts denoted as $\mathbf{P}_k$ and $\mathbf{P}_v \in \mathbb{R}^{l \times d}$, integrated into the key $\mathbf{H}_k$ and values $\mathbf{H}_v$ within the LLaMA model, where $\mathbf{H}_q$ represents the query, the attention module is adapted as follows:
\begin{equation}
    \mathbf{H}_i = \textnormal{Attention}(\mathbf{H}_q, [\mathbf{P}_k; \mathbf{H}_k], [\mathbf{P}_v; \mathbf{H}_v])
\end{equation}
\begin{equation}
    \textnormal{MSA} = \textnormal{Concat}(\mathbf{H}_1,...,\mathbf{H}_m) \mathbf{W}_o
\end{equation}
where $[;]$ denotes concatenation, $\mathbf{W}_o$ is projection matrices, $\mathbf{H}_i$ denotes $i$-th head and $m$ is the number of total heads.  
In this paper, we adopt the complementary learning principle~\cite{Dualprompt}, incorporating learnable General prompts $\mathbf{P}_g$ (G-Prompt) and Expert prompts $\mathbf{P}_e$ (E-Prompt) into the first $j$ layers of the LLaMA model, where the G-Prompt is applied to the first $i$ layers to capture task-invariant knowledge, whereas the E-Prompt is applied to the subsequent layers from $i+1$ to $j$ for task-specific knowledge adaptation. Through this prompting approach, we effectively train a small number of parameters while retaining the knowledge of existing tasks, all without the need for external memory.

\smallskip
\noindent\textbf{Overall Architecture.} Our proposed method, termed collaborative prompting (ColPro) for continual VideoQA, is illustrated in Figure~\ref{fig:overview}. Leveraging LLM-based VideoQA models as a foundation, our goal is to establish a cohesive set of collaborative and interactive prompts. This approach aims to mitigate the issue of catastrophic forgetting often associated with straightforward sequential fine-tuning methods.

Each training task set consists of video $V^t$, question $Q^t$, and answers $A^t$ in pairs. We extract a sequence of visual tokens $\mathbf{V} = \{\mathbf{v}_1, \ldots, \mathbf{v}_{N_v}\} \in \mathbb{R}^{N_v \times D}$ from the raw video using a frozen visual encoder~\cite{clip}, and utilize a tokenizer to process the raw question and answer into tokens, i.e., $\mathbf{Q} = \{\mathbf{q}_1, \ldots, \mathbf{q}_{N_q}\} \in \mathbb{R}^{N_q \times D}$ and $\mathbf{A} = \{\mathbf{a}_1, \ldots, \mathbf{a}_{N_a}\} \in \mathbb{R}^{N_a \times D}$, where $N_v$, $N_q$ and $N_a$ denote the number of video frames, lengths of question and answer tokens, respectively. \underline{During the training stage}, the task-specific token sequences $\mathbf{q}^t$, $\mathbf{v}^t$, and $\mathbf{a}^t$ are concatenated and inputted into LLaMA along with an additional prompts $\mathbf{P}$ = $<$$\mathbf{P}_e, \mathbf{P}_g$$>$ = $\{ \mathbf{p}_1, \ldots, \mathbf{p}_{N_p}\} \in \mathbb{R}^{N_p \times D}$, where $N_p$ denotes length of prompts. This setup allows the output feature to be calculated as follows:
\begin{equation}
{\mathbf{X}^t, \mathbf{P}} =\theta(<\mathbf{Q}^t, \mathbf{V}^t, \mathbf{A}^t>, \mathbf{P}),
\end{equation}
where $\mathbf{X}^t$ = $<$$\mathbf{X}_q^t, \mathbf{X}_v^t, \mathbf{X}_a^t$$>$ denotes the sequence of output features for question, video and answer for task $t$. 
In our framework, $\mathbf{P}_g$ is trained alone using the global cross-entropy loss similar to existing methods~\cite{Dualprompt, l2p}, while we focus on optimizing the E-Prompt $\mathbf{P}_e$ to effectively capture and preserve task-specific knowledge, thereby reducing catastrophic forgetting. \underline{During the inference stage}, LLaMA takes $\mathbf{V}^t$, $\mathbf{Q}^t$, and the learned prompts $\mathbf{P}$ to predict task-specific answers.

\subsection{Collaborative Prompting}
We systematically explore continual learning, focusing on integrating multimodal distributions into a unified set of prompts. This approach provides a comprehensive framework for continuous improvement in LLM-based VideoQA. Our methodology includes collaboratively incorporating prompts designed for task-specific question constraints, visual temporal awareness, and knowledge acquisition.

\smallskip
\noindent\textbf{Task-specific Question Constraint Prompting (TQCP).} TQCP extracts question-specific knowledge from learned prompt representations, enhancing task awareness during the inference stage. Different from existing methods~\cite{Dualprompt, l2p} that rely on a known task identity to select and train specific sets of prompts alongside the classifier, we directly utilize the question type to guide the learning of a single set of prompts for question awareness. Drawing inspiration from a recent negative label guided algorithm~\cite{NegGuided, negpro}, we enable $\mathbf{P}_e$ to be positively correlated with the current task-specific question type (e.g., how many) and negatively correlated with negative question samples (e.g., negative question types: what, where, etc). This facilitates task-type awareness and links input features to E-prompts during the inference stage. To achieve this, we optimize $\mathbf{X}^t_q$ and $\mathbf{P}_e$ with question generation loss and negative questions ($\mathbf{Q}^-$) guided loss, which allows the given prompt to learn question-specific representation for the current task. It can be formulated as follows:
\begin{equation} \label{eq:lq}
    \mathcal{L}_q =  - (\mathcal{L}_q^{gen} + \mathcal{L}_q^{neg})
\end{equation}
\begin{equation}
    \mathcal{L}_q^{gen} 
    = \sum^{N_q-1}_{n=0}\textnormal{log} P(\mathbf{q}_{n+1}^*|\mathbf{V}, \mathbf{A}, \mathbf{P}, \mathbf{q}^+ \le n, \mathbf{Q}^-),
\end{equation}
{\scriptsize
\begin{equation}
    \mathcal{L}_q^{neg} 
    = \frac{1}{\mathcal{B}}\sum_{i \in \mathcal{B}}
    \left(\frac{e^{sim(\mathbf{P}_e, \mathbf{Q}^+_i)/\tau}}{\sum_{j\in \mathcal{B}}(e^{sim(\mathbf{P}_e, \mathbf{Q}^+_j)/\tau} + e^{sim(\mathbf{P}_e, \mathbf{Q}^-_j)/\tau})}\right),
\end{equation}
}
where $P(\mathbf{q}_{n+1}^*)$ = $\textnormal{Softmax}(\textnormal{Linear}([\mathbf{X}_q ; \mathbf{P}_e]))$ for task $t$, and $[;]$ denotes concatenation. $\mathbf{P}$ = $<$$\mathbf{P}_e, \mathbf{P}_g$$>$ and $\tau$ is a temperature parameter. We employ cross-entropy loss $\mathcal{L}_q^{gen}$ locally to generate task-specific questions based on learned $\mathbf{X}_q$ and $\mathbf{P}_e$. $\mathcal{L}_q^{neg}$ to correlate the given question $\mathbf{Q}^+$ with $\mathbf{P}_e$, where $sim(\cdot,\cdot)$ computes the cosine similarities between the $\mathbf{P}_e$ and the $i$-th positive question $\mathbf{Q}^+_i$ (resp. $j$-th negative question $\mathbf{Q}^-_j$) samples in the batch $\mathcal{B}$.

\smallskip
\noindent\textbf{Visual Temporal Awareness Prompting (VTAP).} VTAP aims to bridge the gap between video features and the LLM, allowing E-prompts to incorporate visual information with temporal dynamics. This improves the video understanding abilities of the LLM and enhances its answer prediction capabilities with given questions and videos. However, modeling both the visual content of videos and their temporal dynamics simultaneously presents a challenge. To overcome this, we guide the E-prompt in learning video with temporal dynamics by using the question and answer choices as prior knowledge and leverage the autoregressive sequential abilities of the LLM to model and predict the order of video frames based on preceding frames. Furthermore, we distill video information extracted from an image encoder~\cite{clip} into an E-prompt~\cite{PanDa, blip2}, enabling the LLM to understand visual features. In this work, we use contrastive loss ($\mathcal{L}_q^{con}$) to facilitate this process, which is formulated as follows:
\begin{equation} \label{eq:lv}
    \mathcal{L}_v =  - (\mathcal{L}_v^{dyn} + \mathcal{L}_v^{con})
\end{equation}
\begin{equation}
\begin{split}
    \mathcal{L}_v^{dyn} 
    = \sum^{N_v-1}_{n=0}\textnormal{log} P(\mathbf{v}_{n+1}^*|\mathbf{Q}, \mathbf{A}, \mathbf{P}, \mathbf{v} \le n),
\end{split}
\end{equation}
\begin{equation}
\begin{split}
    \mathcal{L}_v^{con} 
    = \frac{1}{\mathcal{B}}\sum_{i \in \mathcal{B}}\textnormal{log}\left(\frac{e^{sim(\mathbf{P}_{e}, \mathbf{V}_i)}/\tau}{\sum_{j \in \mathcal{B}}e^{sim(\mathbf{P}_{e}, \mathbf{V}_j)}/\tau}\right),
\end{split}
\end{equation}
where $P(\mathbf{v}_{n+1}^*) = \textnormal{Softmax}(\textnormal{Linear}([\mathbf{X}_v ; \mathbf{P}_e]))$ for task $t$, $\mathcal{L}_v^{dyn}$ is the optimization function for video temporal dynamic modelling, and  $sim(\cdot,\cdot)$ computes the cosine similarities between the $\mathbf{P}_e$ and the $i$-th video $\mathbf{V}_i$ (resp. $j$-th video $\mathbf{V}_j$) in the batch $\mathcal{B}$ for current task.

\smallskip
\noindent\textbf{Knowledge Acquisition Prompting (KAP).} 
KAP injects task-specific multimodal information from the question and video into the E-prompts to accurately predict answers for the current task.
To achieve this, at the training stage, $\mathbf{P}_e$ leverages the autoregressive abilities of the LLM to encapsulate the task-specific context information of  $\mathbf{V}$, $\mathbf{Q}$, and all answer choices $\mathbf{A}$ as prior knowledge to predict the specific answer.
The objective function is formulated as:
\begin{equation} \label{eq:la}
\begin{split}
    \mathcal{L}_a 
    = &-\sum^{N_a-1}_{n=0}\textnormal{log} P(\mathbf{a}_{n+1}^*|\mathbf{Q}, \mathbf{V}, \mathbf{P}, \mathbf{a} \le n),
\end{split}
\end{equation}
where $P(\mathbf{a}_{n+1}^*)$ = $\textnormal{Softmax}(\textnormal{Linear}([\mathbf{X}_a; \mathbf{P}_e]))$. 
At the inference phase, the continual VideoQA model predicts the task-specific answer with $\mathbf{V}$, $\mathbf{Q}$, $\mathbf{P}$ as:
\begin{equation}
    \Bar{a} = \arg\max_{a \in \mathcal{A}^t} P(\mathbf{a} | \mathbf{V}, \mathbf{Q}, \mathbf{\mathbf{P}}),
\end{equation}
where $\mathcal{A}^t$ is a set of answer choice for task $t$.

\section{Experiments}


\subsection{Datasets}\label{sec:4.1}


We use the multi-choice NExT-QA dataset \cite{Next-qa}, which includes various types of questions. These include causal questions, such as why (CW) and how (CH), that ask for the intentions or reasons behind earlier actions; temporal questions, which determine the relationships between actions like what are (TC), what did (TN), and what was (TP); and descriptive questions, like how many (DC), where (DL) and other types of question (DO), which focus on visible contents such as places and attributes. We split \cite{Symbolic_Replay, NVQACL} the NExT-QA dataset into eight distinct tasks based on question types in the NExT-QA dataset.
In CL, the order of task learning impacts the learning outcome. Therefore, we conducted experiments and set our tests in the sequence that resulted in the highest forgetting rate (suffers more in catastrophic forgetting) using the baseline method. The sequence of the training order follows this sequence: $<$TP, CW, DC, TC, DL, DO, TN, CH$>$.
DramaQA dataset \cite{drama-qa} features a video story understanding with hierarchical difficulty levels. We split it into 5 distinct tasks according to the question types, and experimentally chose the order that has the worst forgetting result to be our learning order for the CL VideoQA task. The order of CL learning follows $<$what, who, where, how, why$>$. Similar to the existing VideoQA methods~\cite{Self-chained-VQA}, we report the performance of the validation set.


\subsection{Evaluation Metrics} 


We evaluate methods using two metrics: the average final accuracy (Avg. Acc), where higher values are better and represent the final accuracy averaged over N tasks for all previous classes, and the average forgetting (Avg. Fog) that is widely used in existing works \cite{l2p, Dualprompt, Decouple_vqa}, where the lower the better which indicates the tasks experienced less forgetting averaged over N tasks.

\begin{table}[t!]
  \centering
  \caption{The results on the NExT-QA dataset which are divided into 8 tasks, where the Avg. Acc denotes average accuracy across tasks and Avg. Fog is the average forgetting rate. The symbols $\uparrow$ and $\downarrow$ indicate whether a higher or lower value is preferable for a given metric, respectively.}
    \begin{tabular}{l|cc}
    \bottomrule
    Method & \multicolumn{1}{l}{Avg. Acc ($\uparrow$)} & \multicolumn{1}{l}{Avg. Fog ($\downarrow$)} \\
    \hline
    LLaMA &   46.58    & 13.83 \\
    L2p   &   48.82    & 12.25 \\
    DualPrompt &   50.62    & 11.74 \\
    LAE   &   49.38    & 11.47 \\
    \hline
    L2p+  &   52.26    &  11.61\\
    DualPrompt+ &  53.97     & 10.26 \\
    LAE+  &    53.75   & 9.74 \\
    DAM &    53.88   & 9.99 \\
    ProgPrompt &    53.95   & 10.69 \\
    \hline
    ColPro  &   \textbf{55.14}    & \textbf{7.43} \\
    \toprule
    \end{tabular}%
  \label{tab:1} 
\end{table}%


\subsection{Implementation Details}


We train CL VideoQA for five epochs on both datasets with a batch size
of 8 for the NExT-QA dataset (4 for DramaQA) with the gradient accumulation technique. We fine-tuned the LLaMA-7B model in this paper. AdamW optimizer is used with $\beta$ = (0.9, 0.95). We search learning rate and weight decay in [0.05, 0.1] and [0.15, 0.25], respectively. The number of video frames V is set to 10. Each frame is resized by 224 × 224 and fed into CLIP VIT-L/14 to extract frame features. The total sequence length of the concatenated visual, question, and answer tokens is 128 for NExT-QA and 280 for DramaQA. Temperature parameter $\tau$ is set to 1.  The prompt tokens are empirically set to 10 for $\mathbf{p}_k$, $\mathbf{p}_v$. The positing of G-prompt and E-prompt are set to 0-6 and 7-18 LLaMA layers, respectively, for the best performance. The prompts are not attached to the remaining LLaMA layers.



\subsection{Comparison with Continual Learning Methods}

\begin{table}[t!]
  \centering
  \caption{The results on the DramaQA dataset which are divided into 5 tasks.}
    \begin{tabular}{l|cc}
    \bottomrule
    Method & \multicolumn{1}{l}{Avg. Acc ($\uparrow$)} & \multicolumn{1}{l}{Avg. Fog ($\downarrow$)} \\
    \hline
    LLaMA &   60.99    & 24.39 \\
    L2p   &   62.50    & 20.67 \\
    DualPrompt &   65.89    & 17.93 \\
    LAE   &   65.82    & 17.35 \\
    \hline
    L2p+  &   66.75    &  16.73\\
    DualPrompt+ &  67.44     & 15.09 \\
    LAE+  &    67.03   & 14.82 \\
    DAM   &    67.37   & 15.19 \\
    ProgPrompt &    67.92   & 14.95 \\
    \hline
    ColPro  &   \textbf{71.24}    & \textbf{12.64} \\
    \toprule
    \end{tabular}%
  \label{tab:2} 
\end{table}%

Table \ref{tab:1} compares the performance of the Collaborative Prompting (ColPro) on the split NExT-QA benchmark with existing CL approaches, including the fine-tuned LLaMA~\cite{llama} with addtional projection layer, L2P~\cite{l2p}, DualPrompt~\cite{Dualprompt}, ProPrompt~\cite{progprompt}, DAM~\cite{DAM}, and LAE~\cite{LAE}. Additionally, we report deep CL implementations of the L2P+, DualPrompt+, and LAE+ methods, which activate more layers of LLaMA for CL tasks by applying prompts to 18 layers. In the comparisons, our proposed ColPro achieved better average prediction accuracy and significantly lower average forgetting compared to existing methods. This improvement in average forgetting can be attributed to the fact that the ColPro method experiences less forgetting and allows better forward transfer of different tasks, which is beneficial in CL VideoQA.
Similarly, we compare the performance of the ColPro on the split DramaQA benchmark with existing CL approaches in Table \ref{tab:2}, further validating the effectiveness of our proposed method in addressing catastrophic forgetting issues. These tables indicate that the models experience catastrophic forgetting, with the Avg. Fog score up to 24\%. This underscores the need to address catastrophic forgetting in video QA, and we have minimized the forgetting with ColPro.

\begin{table}[t!]\setlength{\tabcolsep}{1.5pt}
  \centering
  \caption{The results on both NExT-QA and DramaQA datasets with PEFT and our methods.}
    \begin{tabular}{l|c|cc}
    \bottomrule
          \multicolumn{1}{c|}{Methods} & \multicolumn{1}{c|}{Dataset} & \multicolumn{1}{c}{Avg. Acc ($\uparrow$)} & \multicolumn{1}{c}{Avg. Fog ($\downarrow$)} \\
          \hline
    Prefix & \multirow{4}[1]{*}{NExT-QA} &   47.76    & 13.21 \\
    Lora  &       &   52.00    &  11.07 \\
    L-Adaptor &       &  51.83     &  12.42 \\
    Ours  &       &    \textbf{55.14}    & \textbf{7.43}  \\
    \hline
    Prefix & \multirow{4}[2]{*}{DramaQA} &   60.93    & 21.18 \\
    Lora  &       &   62.11    &  19.73  \\
    L-Adaptor &       &   61.50    & 19.77 \\
    ColPro  &       &  \textbf{71.24}    & \textbf{12.64} \\
    \toprule
    \end{tabular}%
  \label{tab:3} 
\end{table}%

\begin{figure}[t]
  \centerline{\includegraphics[width=8cm]{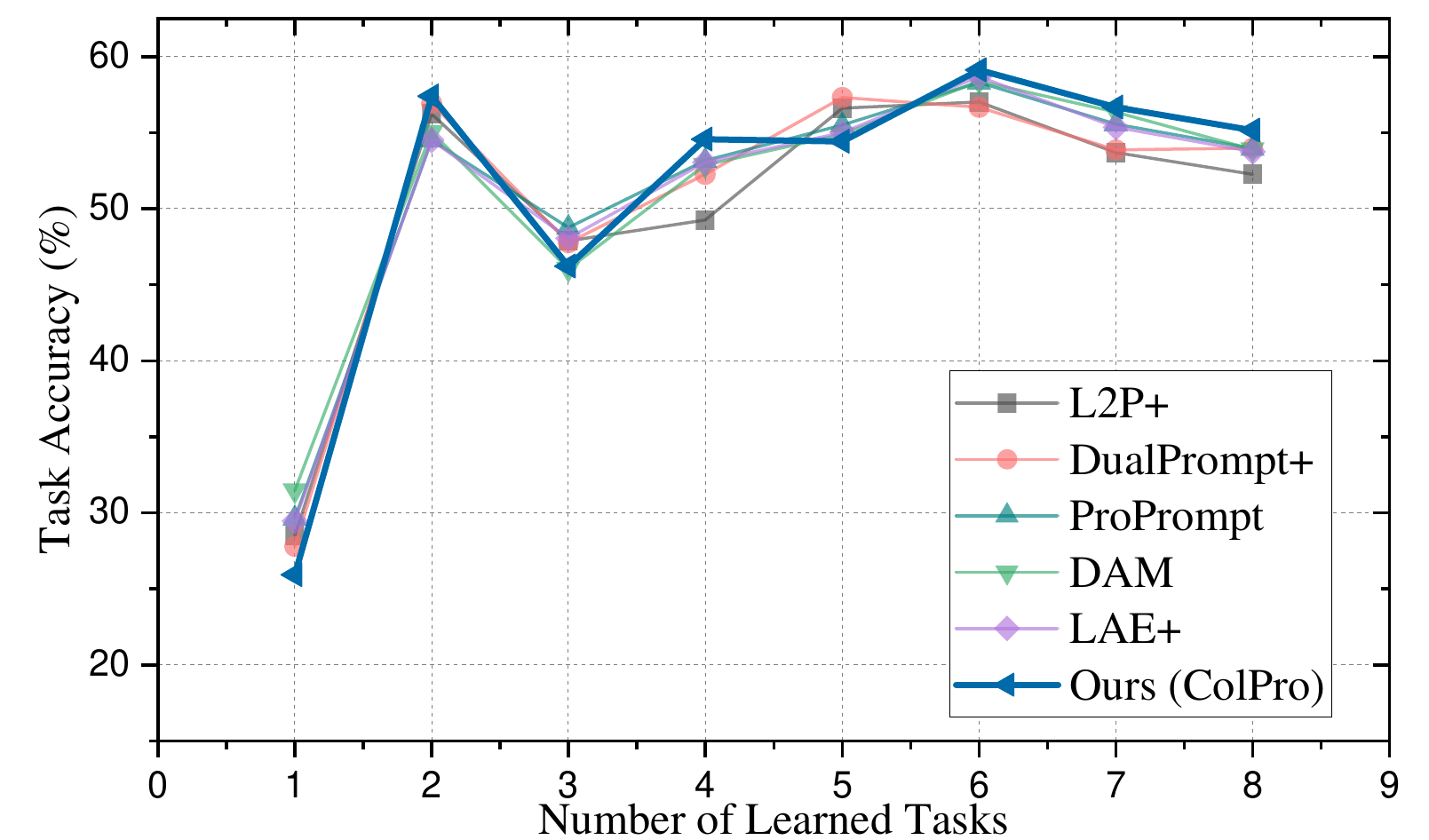}}
  \caption{The results of the average accuracy for each task, which following the training order within the CL setting.}
  \label{fig:3} 
\end{figure}


\subsection{Comparison with Parameter-Efficient Fine Tuning Methods}


Table~\ref{tab:3} demonstrates the performance of ColPro with other Parameter-Efficient Fine-Tuning (PEFT) methods that can be used to address catastrophic forgetting in CL settings, such as LLaMA-Adapter (L-Adapter)~\cite{Llama-adapter}, Lora~\cite{lora}, and Prefix~\cite{Prefix-tuning}. Our proposed method shows a significant improvement in minimizing forgetting compared to our baseline LLaMA-Adapter method, as evidenced by a lower Avg. Fog and a higher Avg. ACC when evaluated with the NExT-QA and DramaQA datasets. ColPro also outperforms Lora and Prefix, demonstrating the effectiveness of our specially designed strategy for LLM-based CL VideoQA settings.

\begin{table}[t!]
  \centering
  \caption{The ablation results for the three proposed multimodal interaction prompting strategies.}
    \begin{tabular}{ccc|cc}
    \bottomrule
    \multicolumn{1}{l}{$\mathcal{L}_a$} & \multicolumn{1}{l}{$\mathcal{L}_q$} & \multicolumn{1}{l|}{$\mathcal{L}_v$} & \multicolumn{1}{l}{Avg. Acc ($\uparrow$)} & \multicolumn{1}{l}{Avg. Fog ($\downarrow$)} \\
    \hline
       \cmark   &   \xmark    &  \xmark     &   52.60    & 10.62 \\
       \cmark   &   \cmark    &  \xmark     &    53.09   & 9.09 \\
       \cmark   &   \xmark    &  \cmark     &   54.38    &  10.27 \\
       \cmark   &   \cmark    &  \cmark    &  \textbf{55.14}    & \textbf{7.43}  \\
    \toprule
    \end{tabular}%
  \label{tab:4} 
\end{table}%

\begin{table}[t!]
\setlength{\tabcolsep}{3.2pt}
  \centering
  \caption{The ablation results for different prompting optimization functions.}
    \begin{tabular}{ccc|cc}
    \bottomrule
    \multicolumn{1}{l}{$\mathcal{L}_q^{neg}$} & \multicolumn{1}{l}{$\mathcal{L}_v^{dyn}$} & \multicolumn{1}{l|}{$\mathcal{L}_v^{con}$} & \multicolumn{1}{l}{Avg. Acc ($\uparrow$)} & \multicolumn{1}{l}{Avg. Fog ($\downarrow$)} \\
    \hline
       \xmark   &   \xmark    &  \xmark  &   52.87    & 10.15  \\
       \cmark   &   \xmark    &  \xmark     &   53.09    & 9.09 \\
       \cmark   &   \cmark    &  \xmark     &   52.80   & 9.95 \\
       \cmark   &   \xmark    &  \cmark     &   54.20    & 8.71 \\
       \cmark   &   \cmark    &  \cmark    &   \textbf{55.14}    & \textbf{7.43}  \\
    \toprule
    \end{tabular}%
  \label{tab:5} 
\end{table}%


\subsection{Task-by-Task Average Accuracy}


Continuous learning (CL) in real-world scenarios is an ongoing process, making the performance of each learning phase critical for the VideoQA model. Consequently, we plotted the task-by-task continual learning average accuracy in Figures~\ref{fig:3} for the NeXT-QA dataset with respect to the order of training. We accumulated with the previous tasks, and tested on the current task to report the average accuracy. 
Our results show that ColPro achieves better performance in most of the perdition accuracy than existing methods in the learning phase.


\subsection{Ablation Study}


\textbf{The Effectiveness of Multimodal Prompting.} Our proposed method includes three primary multimodal interaction prompting strategies: question constraint (TQCP), visual temporal alignment (VTAP), and knowledge acquisition (KAP). Each strategy is optimized with its respective optimization function, $\mathcal{L}_q$, $\mathcal{L}_v$, and $\mathcal{L}_a$. We performed ablation studies on these strategies and reported the results in Table~\ref{tab:4}. A tick indicates that the respective prompting strategies was used during model training, while a cross means not included. Notably, $\mathcal{L}_a$ is always included to optimize the model with the ground truth answer.
We can observe that the inclusion of each optimization function significantly impacts the model's performance. Specifically, models trained with all three optimization functions consistently achieve higher accuracy and lower forgetting.
The ablation results demonstrated show that incorporating VTAP enhances the accuracy of the LLM, while utilizing TQCP helps the model suffer less from forgetting. 
This underscores the importance of question constraint and visual temporal alignment prompting, which help the LLM gain awareness of the task type to reduce catastrophic forgetting and understand the visual information with temporal dynamics for better answer reasoning.
The combination of three strategies is crucial for optimizing multimodal interaction in CL VideoQA scenarios.

Furthermore, in Table~\ref{tab:5}, we break down each prompting optimization to evaluate the effectiveness of using negative contrast ($\mathcal{L}_q^{neg}$), visual distillation ($\mathcal{L}_v^{con}$), and temporal dynamic understanding ($\mathcal{L}_v^{dyn}$) techniques. The optimization functions $\mathcal{L}_a$ and $\mathcal{L}_q^{gen}$ are always included to optimize the model with the ground truth answer and question during training.
We can observe that $\mathcal{L}_q^{neg}$ tends to reduce forgetting in large margin, indicating that the negative contrast technique allows question constraint prompts to gain task-specific knowledge, making the model better aware of the task type. 
Furthermore, we can see that the stand along temporal dynamic ($\mathcal{L}_v^{dyn}$) does not benefit the model, but the visual distillation ($\mathcal{L}_v^{con}$) is able to improve the average accuracy. The method achieves excellent performance when the model the combination of $\mathcal{L}_v^{dyn}$ and $\mathcal{L}_v^{con}$. This emphasis the importance of the proposed temporal dynamic understanding that bridges LLM with video information for VideoQA to achieve better performance.

\begin{table}[t!]
  \centering
  \caption{The ablation results for prompt positioning in LLaMA layers.}
    \begin{tabular}{cc|cc}
    \bottomrule
    \multicolumn{1}{l}{$\mathbf{p}_g^{end}$} & \multicolumn{1}{l|}{$\mathbf{p}_e^{end}$} & \multicolumn{1}{l}{Avg. Acc ($\uparrow$)} & \multicolumn{1}{l}{Avg. Fog ($\downarrow$)} \\
    \hline
      18    &    0   &   52.53    & 10.45 \\
       0   &    18   &   53.66    & 8.81 \\
       6   &    18   &   53.74    & 9.01 \\
       10   &    18   &   54.78    & 8.11 \\
       8   &    16   &  54.45  &  8.59 \\
       8   &    18   &  \textbf{55.14}  & \textbf{7.43} \\
       8   &    20   &  54.75    &  9.02 \\
    \toprule
    \end{tabular}%
  \label{tab:6} 
\end{table}%

\begin{table}[t!]
\setlength{\tabcolsep}{8pt}
  \centering
  \caption{The ablation results for using different lengths of the prompt in the model.}
    \begin{tabular}{c|cc}
    \bottomrule
    \multicolumn{1}{l|}{Lenght of $\mathbf{p}$} & \multicolumn{1}{l}{Avg. Acc ($\uparrow$)} & \multicolumn{1}{l}{Avg. Fog ($\downarrow$)} \\
    \hline
        5   &   53.91    & 8.91 \\
       10   &   \textbf{55.14}    & \textbf{7.43}  \\
      15   &    54.55    & 8.35 \\
      20   &    53.77   & 9.63 \\
    \toprule
    \end{tabular}%
  \label{tab:7} 
\end{table}%

\smallskip
\noindent\textbf{Number of layers for G-Prompts and E-Prompts.} In our method, we utilize both E-prompt and G-prompt. In Table~\ref{tab:6}, we empirically evaluate the effectiveness of placing the G-prompt and E-prompt within the LLaMA layers, following existing methods, to achieve the best performance. $\mathbf{p}_g$ and $\mathbf{p}_e$ denote the last layer attached to LLaMA-7B, the pre-trained network with 32 layers. The table shows that performance drops when either the $\mathbf{p}_g^{end}$ = 0 or E-prompt $\mathbf{p}_e^{end}$ = 0 is excluded.
We found that the optimal performance was achieved by attaching the G-Prompt from layers 0 to 8 and the E-Prompt from layers 9 to 18.

\smallskip
\noindent\textbf{Number of prompt.} For prompt-based learning, the length of the prompt is a crucial parameter that can significantly affect learning performance \cite{ LAE, l2p}. Table~\ref{tab:7} illustrates the impact of varying prompt lengths on model accuracy and forgetting rates.
Our results indicate that shorter prompts may not provide sufficient context for the model, leading to lower accuracy and higher forgetting. Conversely, excessively long prompts can introduce noise and unnecessary information, which also negatively impacts performance in our experiment. We found an optimal prompt length of 10 to balance the amount of information provided to the model and maintain high performance.


\section{Conclusion}


In this work, we explore the novel problem of VideoQA, which efficiently fine-tunes the LLM to answer new questions with video in a continual learning context. We propose Collaborative Prompting (ColPro) to integrate textual question context, visual content, and video temporal dynamics in each learning phase, facilitating knowledge transfer to future tasks while minimizing catastrophic forgetting. 
We achieves state-of-the-art results on split NExT-QA and DramaQA datasets.

\section{Limitations}
We propose the efficient Collaborative Prompting (ColPro), which integrates task-specific question constraint prompting, knowledge acquisition prompting, and visual temporal awareness prompting with a large language model (LLM) to enhance the performance of continual VideoQA. However, catastrophic forgetting remains high for the DramaQA dataset using our method, indicating a substantial decline in performance for VideoQA prediction when using LLaMA-7B. Furthermore, we did not experiment with larger models (e.g., 33B LLM) due to memory constraints, which limits our ability to explore catastrophic forgetting that may arise when fine-tuning other LLMs for CL VideoQA.

\noindent\textbf{Acknowledgments:} This research/project is supported by the National Research Foundation, Singapore under its AI Singapore Programme (AISG Award No: AISG-PhD/2021-08-024[T]). Any opinions, findings and conclusions or recommendations expressed in this material are those of the author(s) and do not reflect the views of National Research Foundation, Singapore.




\bibliography{custom}
\bibliographystyle{acl_natbib}

\appendix

\section{Appendix}
\label{sec:appendix}


\subsection{Critical Continual Learning Order}
In Table \ref{tab:8}, we outline the sequence of continual learning tasks in VideoQA, enabling us to identify and select the most critical tasks that are particularly susceptible to catastrophic forgetting. By understanding which learning order is most affected by this phenomenon, we can prioritize and implement targeted strategies to mitigate forgetting, thereby enhancing the overall robustness and effectiveness of the continual learning system. We use baseline model \cite{llama} with additional linear layer for this experiment. We can see that the learning sequences $<$TP, CW, DC, TC, DL, DO, TN, CH$>$ have higher Avg. Fog for NExT-QA dataset.

\begin{table}[h]
\setlength{\tabcolsep}{2pt}
  \centering
  \caption{The results of the task learning sequences for continual learning for the NExT-QA datasets.}
    \begin{tabular}{c|c}
    \bottomrule
    \multicolumn{1}{l|}{Orders} & \multicolumn{1}{l}{Avg. Fog ($\downarrow$)} \\
    \hline
    CH, DL, TP, TC, DC, DO, TN, CW  & 7.26  \\
    TP, TN, CH, TC, DL, DO, CW, DC    & 8.41 \\
    DO, CW, DC, CH, TP, TC, TN, DL    & 9.83 \\
    CW, DO, TN, DL, TC, TP, DC, CH    & 11.86 \\
    TP, CW, DC, TC, DL, DO, TN, CH    & 13.83 \\
    \toprule
    \end{tabular}%
  \label{tab:8} 
\end{table}%

\begin{figure*}[ht!]
 \centerline{\includegraphics[width=16cm]{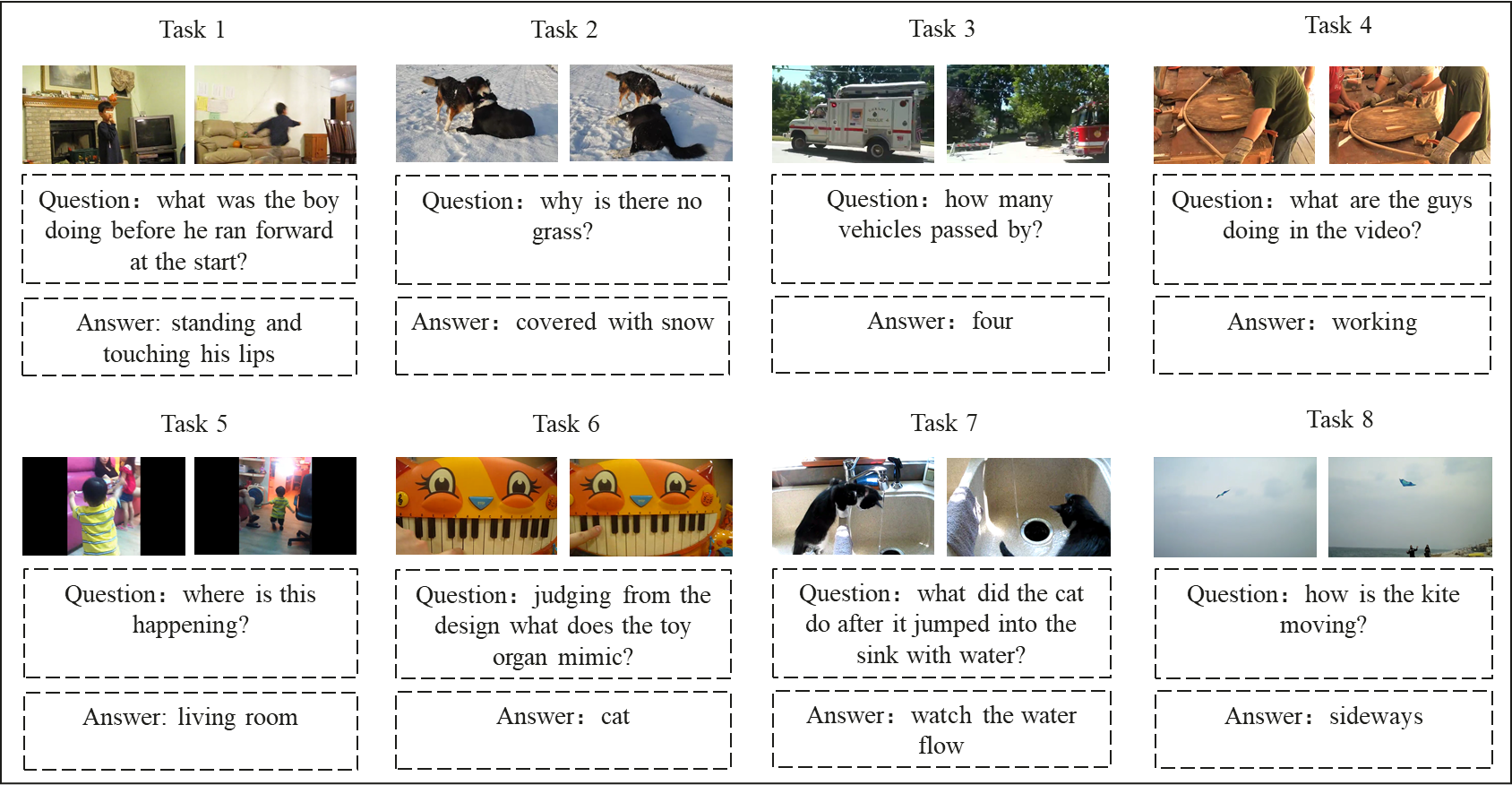}}
  \caption {The video examples with their corresponding questions and answers for each task.}
  \label{fig:tasks} 
\end{figure*}

\subsection{Learning Parameter Analysis with Full Model Fine-tuning}
 Since most parameters are fixed at the inference stage, the performance of a fine-tuned prompt-based model may be worse than that of a fully fine-tuned model for each specific task. However, during the training stage, fine-tuning the entire LLM incurs high computational costs. Here, we provided an analysis of this aspect to better understand the trade-offs between the effectiveness and computation cost of these two approaches using the score we get from the DramaQA dataset as an example. According to the training parameters indicated in LLaMa \cite{llama} and Lora \cite{lora}, we can assume that to fully fine-tune an LLM requires more than 500M parameters, whereas our prompt-based method only requires around 33.5M parameters. Although the Avg. Acc (assumed to be >71.24) of full LLaMA fine-turning may be higher than our score, but it requires a much higher computation cost. Our method can efficiently and effectively fine-tune LLaMA-7B model for CL in VideoQA using a single 24GB GPU. Furthermore, ColPro achieved better performance compared to existing prompt-based methods, as illustrated in Tables \ref{tab:1} and \ref{tab:2}. It's worth noting that due to limited computational power, we were unable to provide the results for full LLaMA fine-tuning.

\begin{table}[t!]\setlength{\tabcolsep}{8pt}
    \centering
    \caption{Analysis of Model Parameters and Average Accuracy}
    \label{tab:model_comparison}
    \begin{tabular}{@{}lrr@{}}
        \toprule
        \textbf{Model} & \textbf{Parameters} & \textbf{Avg. Acc} \\
        \midrule
        LLaMA fine-tuning & $>500\,\text{M}$ & $>71.24$ \\
        Prompt-tuning & $33.5\,\text{M}$ & $71.24$ \\
        \bottomrule
    \end{tabular}
\end{table}


\subsection{Continual Learning Setting and Examples}
In this paper, we split the dataset towards the function-incremental setting of continuous learning, similar to existing CL ImageQA works \cite{Symbolic_Replay, Decouple_vqa}, to better evaluate the CL VideoQA task. We split the dataset according to different functions. For instance, we split NExT-QA into causal reasoning function, which includes logic understanding of asking why (CW) and how (CH), temporal reasoning function that involves the relationship understanding of objects or attributes recognition in what are (TC), what did (TN), and what was (TP), and descriptive reasoning function encompasses knowledge understanding of how many (DC), where (DL), and other types of questions (DO), as illustrated in Section \ref{sec:4.1}. Similar for DramaQA, we split the dataset according to the function of each question type.  The raw video examples for CL VideoQA with their corresponding question type and answer are illustrated in Figure \ref{fig:tasks}. The figure showing the differences between them for NExT-QA \cite{Next-qa} dataset.







\end{document}